\def\CloudEdgeDocumentClassLoaded{}
\documentclass{article}

\def\CloudEdgeCameraReady{}
\def\CloudEdgeArxiv{}
\def\CloudEdgeIncludeAppendix{}
\def\CloudEdgeAppendixPageBreak{}

\def\CloudEdgeAuthors{
Daojie Peng$^{1}$, Fulong Ma$^{1}$, Bingtao Wang$^{2}$,
Sheng Wang$^{3}$, Jun Ma$^{1,3,*}$\\
$^{1}$HKUST(GZ) \quad $^{2}$Shandong University \quad $^{3}$HKUST\\
\texttt{dpeng108@connect.hkust-gz.edu.cn, fmaaf@connect.hkust-gz.edu.cn,}\\
\texttt{wangbt@mail.sdu.edu.cn, swangei@connect.ust.hk, jun.ma@ust.hk}\\
$^{*}$Corresponding author
}
\def\CloudEdgePDFAuthors{Daojie Peng; Fulong Ma; Bingtao Wang; Sheng Wang; Jun Ma}

\ifdefined\CloudEdgeDocumentClassLoaded
\else
\documentclass{article}
\fi
\usepackage{iclr2027_conference,times}

\usepackage[hyphens]{url}
\usepackage{hyperref}
\usepackage{graphicx}
\usepackage{caption}
\usepackage{placeins}
\usepackage{float}
\usepackage{amsmath}
\usepackage{amssymb}
\usepackage{bm}
\usepackage{newfloat}
\usepackage{listings}
\usepackage{booktabs}
\usepackage{multirow}
\usepackage{xcolor}
\usepackage{booktabs}
\usepackage{subcaption}

\def\UrlFont{\rm}
\def\CloudEdgePaperTitle{Latency-Tolerant Cloud-Edge Collaborative Vision-Language-Action Models via Emergent Representational Specialization}
\title{\CloudEdgePaperTitle}

\ifdefined\CloudEdgeCameraReady
\iclrfinalcopy
\author{\CloudEdgeAuthors}
\hypersetup{
    hidelinks,
    pdftitle={\CloudEdgePaperTitle},
    pdfauthor={\CloudEdgePDFAuthors}
}
\fi

\begin{document}

\maketitle

\ifdefined\CloudEdgeArxiv
\lhead{Preprint}
\fi

\begin{abstract}
Deploying billion-parameter Vision-Language-Action (VLA) policies on mobile robots creates a systems conflict: semantic reasoning benefits from cloud GPUs, whereas closed-loop control must respond locally despite network delay and jitter.
Existing hierarchical and asynchronous policies improve throughput, but their slow-path representations can still arrive stale or require explicit scheduling and delay cues.
We introduce CloudEdgeVLA, a cloud-edge policy that treats temporal misalignment as a representation-learning problem.
A cloud VLA encodes delayed observations into slowly varying task features, while a lightweight edge head combines the latest available cloud feature with current local vision.
During training, current and randomly delayed frames are paired with the same current action target in fresh and stale paths. This objective encourages the cloud representation to preserve task-level information while the edge path supplies state-sensitive corrections, driving emergent specialization.
Across four LIBERO suites, CloudEdgeVLA retains 63.8-78.0\% success with a 40-step uniform-delay window, whereas VLASH reaches at most 6.4\% and the evaluated single-path baselines at most 3.0\%.
By removing blocking synchronization from the control loop, the design offers a practical route to scalable VLA deployment in which cloud models can grow while edge computation remains lightweight and responsive. 
\end{abstract}

\ifdefined\CloudEdgeCameraReady
\noindent\textbf{Code and Project Page:} \pdfstartlink attr{/Border[0 0 0]} user{/Subtype/Link/A<</S/URI/URI(https://daojiepeng.github.io/CloudEdgeVLA/)>>}CloudEdgeVLA\pdfendlink
\fi

\section{Introduction}

Vision-Language-Action (VLA) models transfer semantic knowledge from large vision-language backbones to robot control. RT-2, OpenVLA, $\pi_0$, and Octo demonstrate increasingly broad language following and manipulation capabilities~\citep{brohan2023rt2,kim2024openvla,black2024pi0,team2024octo}. Their scale, however, creates a deployment bottleneck: the compute needed for a multi-billion-parameter backbone is difficult to place on a power- and weight-constrained robot, while manipulation still requires a responsive closed loop.

Cloud robotics can move expensive perception and planning to remote accelerators~\citep{kehoe2015cloudrobotics,wan2016cloudrobotics}. This split makes larger policies deployable, but it also exposes the controller to communication delay, jitter, and loss~\citep{chinchali2019offloading}. A feature returned by the cloud describes the scene when its input image was captured, not necessarily the scene in which the action will execute. Blocking until a new feature arrives lowers the control rate; executing with the latest feature avoids blocking but introduces temporal mismatch.

Multi-rate policies provide part of the solution. MResT combines low-frequency global features with high-frequency local sensing~\citep{saxena2023mrest}; DP-VLA and HiRT separate slow semantic reasoning from fast visuomotor control~\citep{han2024dpvla,zhang2025hirt}; and SmolVLA decouples action generation from execution through an asynchronous inference stack~\citep{shukor2025smolvla}. More recent correction and semantic-action decoupling methods explicitly address inference-time staleness~\citep{sendai2025a2c2,yan2026acting}. These systems establish the value of fast-slow execution, but they primarily target inference scheduling, action-chunk correction, or explicit temporal conditioning. The cloud-edge setting additionally requires the representation passed across the network to remain useful when its age is variable and not known in advance.

\begin{figure*}[t]
    \centering
    \includegraphics[width=\textwidth]{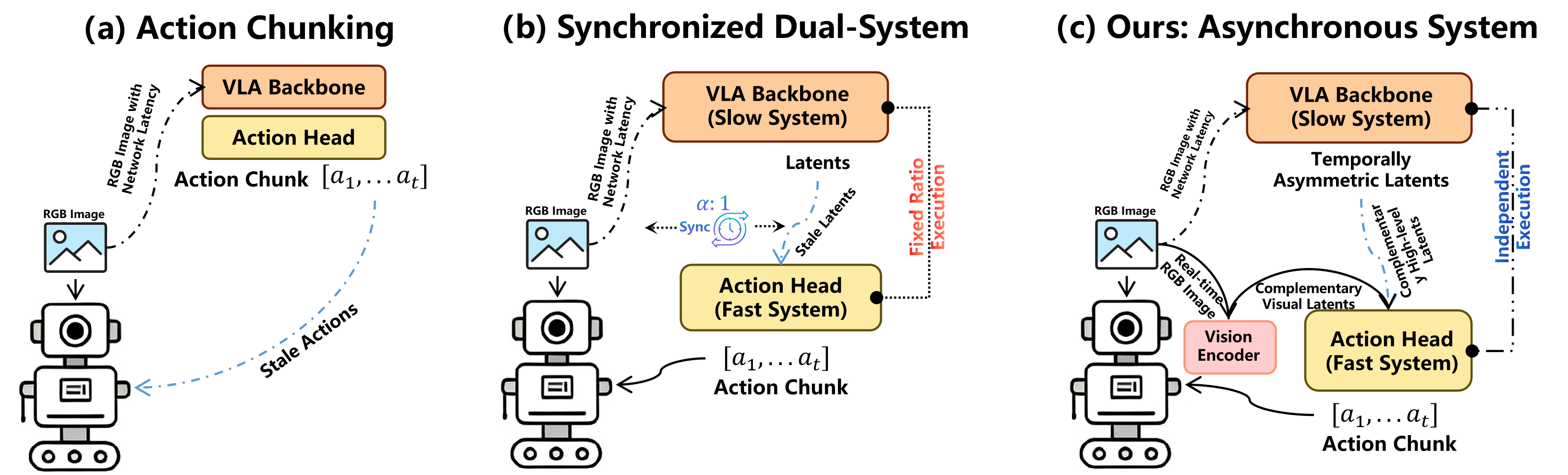}
    \caption{\textbf{Conceptual comparison of dual-system paradigms.}
    (a) Action-chunk re-planning assumes environment stasis during open-loop execution, compounding staleness.
    (b) Synchronized fast-slow pipelines require fixed frequency ratio $\alpha$ and bounded latency--conditions rarely met in networked deployments.
    (c) Our asynchronous paradigm: the cloud produces time-invariant planning features; the edge grounds them with real-time vision without any temporal alignment.
    The edge never blocks on a cloud response and uses the latest available feature.}
    \label{fig:concept}
\end{figure*}

CloudEdgeVLA addresses this requirement by separating information according to temporal role, following the System~1/System~2 analogy~\citep{kahneman2011thinking}. The cloud backbone provides slowly varying task context (what to do), and the edge head uses current local vision for state-sensitive control (how to do it now). 
The edge always consumes the most recently received cloud feature and never waits for a particular update. To make that feature useful across ages, paired-frame dual-path training applies the same current action supervision to cloud features computed from both current and randomly delayed frames. This creates pressure for the cloud path to retain information shared across the pair, while current local vision resolves state details needed at execution time. 
Unlike prior slow‑fast pipelines which pre‑define module roles, our specialization emerges end‑to‑end from training data and loss function, rather than explicit architectural constraints. As illustrated in Figure~\ref{fig:concept}, our edge never stalls for cloud updates; it always consumes the latest arrived feature, even when that feature corresponds to an observation captured many steps ago.

\textbf{Contributions.}
\begin{enumerate}
    \item We formulate asynchronous cloud-edge VLA control around a non-blocking interface: a lightweight Vision-Augmented Action Head fuses the latest cloud feature with current edge vision without requiring clock or frequency alignment.
    \item We propose paired-frame dual-path training: 
    by supervising both fresh and staled cloud features against the identical current‑step action target, our objective induces emergent representational specialization without hand‑crafted invariance regularization. The cloud learns delay‑robust task context, and the edge compensates for instantaneous local state changes.
    \item We conduct systematic experiments on LIBERO benchmarks under controlled uniform network delay, including ablations disentangling the respective robustness contributions of cloud backbone representations and edge‑side correction. We further validate physical feasibility via a small‑scale real‑robot Franka pick‑and‑place pilot study.
\end{enumerate}

\section{Related Work}

\subsection{Vision-Language-Action Models}

RT-2 represents robot actions as language tokens, while OpenVLA provides an open 7B-parameter policy trained on diverse robot data~\citep{brohan2023rt2,kim2024openvla}. Continuous and generative action heads improve control quality and throughput: OpenVLA-OFT combines parallel action-chunk decoding with continuous regression, and $\pi_0$ uses flow matching~\citep{openvlaoft2025,black2024pi0}. Octo and UniVLA explore generalist policies based on transformer and unified multimodal tokenization, respectively~\citep{team2024octo,xie2025univla}. These works establish strong policy backbones; our focus is the systems and learning problem created when backbone features cross a delayed network boundary.

\subsection{Dual-System Robot Architectures}

Action chunking reduces effective planning horizon but can weaken feedback during chunk execution~\citep{karkus2024actionchunk,chi2023diffusionpolicy}. Multi-rate alternatives pair a slow semantic module with a fast local policy: MResT uses different sensing rates, DP-VLA and HiRT couple slow VLM reasoning with faster control, and SmolVLA generates chunks asynchronously~\citep{saxena2023mrest,han2024dpvla,zhang2025hirt,shukor2025smolvla}. Fast-in-Slow embeds fast execution within a slow VLM~\citep{chen2025fastinslow}. Among asynchronous methods, VLASH rolls robot state forward to execution time and evaluates delay on all four LIBERO suites, while A2C2 applies current-observation corrections to stale chunks~\citep{tang2025vlash,sendai2025a2c2}. Semantic-action decoupling instead uses history and time-misalignment training to interpret stale semantics~\citep{yan2026acting}. CloudEdgeVLA differs by sending a learned representation rather than actions across the slow-fast boundary and training it with paired current/delayed frames without delay metadata at inference.

\subsection{Latency Robustness in Control}

Delayed observations have also been studied independently of VLAs. Concurrent control lets a robot act while policy computation proceeds~\citep{xiao2020concurrent}; delayed-observation RL uses history augmentation or delay-resolved state estimates~\citep{wang2024signal}; and world-model methods predict a current latent state from delayed inputs~\citep{karamzade2024worldmodels}. CloudEdgeVLA does not reconstruct the current state. It reserves a direct, current visual path at the edge and trains the cloud feature to supply complementary task context. This design connects delayed-control learning with the established cloud-robotics goal of offloading expensive computation without making the local loop depend on network response time~\citep{kehoe2015cloudrobotics,chinchali2019offloading}.

\section{Method}

\subsection{Problem Formulation}

We formalize the asynchronous cloud-edge VLA deployment problem as follows.
At each environment step $t$:
\begin{itemize}
    \item The \textbf{edge} captures the current observation $o_t$ and sends it to the cloud.
    \item Due to network latency, the cloud receives $o_{t-k}$ (delayed by $k$ steps) and produces planning features $h_{t-k} = f_\theta(o_{t-k}, \ell)$, where $\ell$ is the language instruction and $f_\theta$ is the VLA backbone.
    \item The \textbf{edge} must produce the action $a_t$ using the stale planning features $h_{t-k}$ and its own real-time observation $o_t$.
\end{itemize}
The goal is to learn an action head $g_\phi$ such that $\hat{a}_t = g_\phi(h_{t-k}, v_\psi(o_t))$ achieves high task success despite the temporal mismatch between $h_{t-k}$ and $o_t$, where $v_\psi$ is a local vision encoder.

\begin{figure*}[t]
    \centering
    \includegraphics[width=\textwidth]{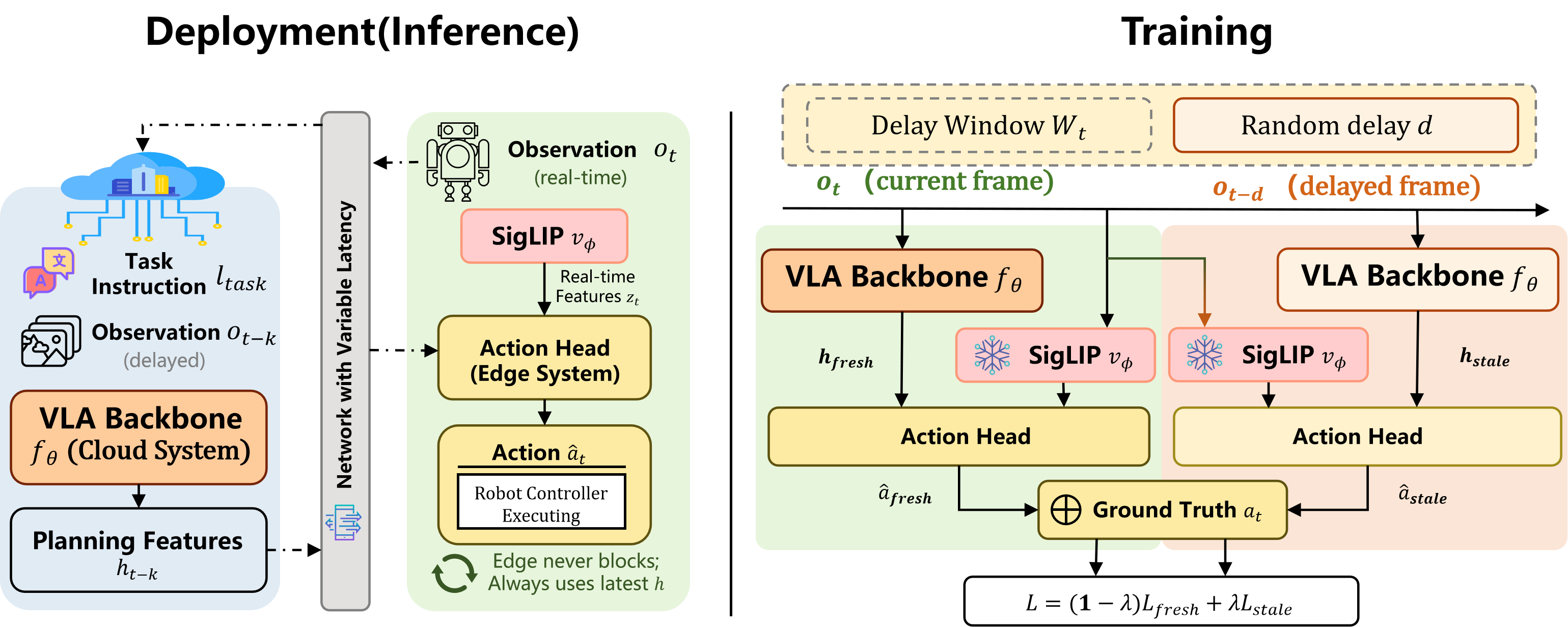}
    \caption{\textbf{System architecture and training pipeline.}
    \textbf{Left:} At deployment time, the cloud VLA backbone produces planning features from a delayed observation, while the edge vision encoder extracts real-time features from the current observation. The action head fuses both to produce actions without any blocking or temporal alignment.
    \textbf{Right:} During training, paired frames from the same episode (current and randomly delayed) are processed through the backbone, producing fresh and stale planning features. Both are fused with the same real-time vision features and supervised toward the same ground-truth action, inducing representational specialization.}
    \label{fig:architecture}
\end{figure*}

\subsection{System Architecture}

Our framework consists of three components operating under a clear temporal asymmetry (see Figure~\ref{fig:architecture}):

\textbf{Cloud-Side VLA Backbone $f_\theta$ (System~2, Latency-Insensitive).}
A large-scale vision-language model that processes visual observations and language instructions to produce high-level planning representations.
Given an observation $o$ and instruction $\ell$, the backbone produces hidden-state representations:
\begin{equation}
    h = f_\theta(o, \ell) \in \mathbb{R}^{L \times D}
\end{equation}
where $L$ is the number of action-token positions (corresponding to an action chunk of length $T$ with $A$ action dimensions, so $L = T \times A$) and $D$ is the hidden dimension.
The backbone is parameterized by LoRA-adapted weights~\citep{hu2022lora} on top of a pretrained vision-language model and runs exclusively on the cloud server.
Critically, the backbone is designed to learn features that encode \emph{what} to do (task goals, manipulation strategy, object semantics) rather than \emph{exactly when} to do it, as motivated in \emph{Emergent Representational Specialization}.

\textbf{Edge-Side Vision Encoder $v_\psi$ (System~1 Perception, Latency-Sensitive).}
A lightweight vision encoder (e.g., SigLIP-Base~\citep{zhai2023siglip}) that runs locally on the robot and extracts real-time visual features from the current observation:
\begin{equation}
    z_t = v_\psi(o_t) \in \mathbb{R}^{D_v}
\end{equation}
The vision encoder is frozen during training to leverage pretrained visual representations.
Crucially, $z_t$ is always computed from the \emph{current} observation $o_t$, making it a \textbf{fresh signal} that captures the instantaneous state of the environment.

\textbf{Edge-Side Action Head $g_\phi$ (System~1 Control, Latency-Sensitive).}
A learnable module that fuses the (potentially stale) cloud planning features with the real-time edge vision features to predict continuous actions:
\begin{equation}
    \hat{a}_t = g_\phi(h_{t-k}, z_t) \in \mathbb{R}^{T \times A}
\end{equation}
The action head first mean-pools the planning features over the action-dimension axis to obtain per-timestep planning embeddings, projects the vision features into the same latent space, concatenates them, and passes the result through a residual MLP to predict the action chunk.
The action head acts as a \textbf{real-time grounding layer}: it translates the cloud's high-level (but potentially stale) directives into precise motor commands using the edge's current visual context.

\subsection{Paired-Frame Dual-Path Training}

The key challenge is that the action head must learn to produce correct actions from \emph{both} fresh and stale planning features.
We achieve this through a paired-frame training strategy that leverages the temporal structure of demonstration trajectories.

\textbf{Paired Frame Extraction.}
During training, each sample provides a window of $W$ consecutive observations from the same episode: $\mathcal{W}_t = \{o_{t-W+1}, \ldots, o_t\}$.
From this window, we construct: (1) \textit{Current frame} $o_t$ (the latest observation) and (2) \textit{Delayed frame} $o_{t-d}$, where $d \sim \text{Uniform}(1, W{-}1)$.

\textbf{Dual Forward Pass.}
Both frames are processed through the cloud backbone:
\begin{equation}
    h^{\text{fresh}} = f_\theta(o_t, \ell), \quad h^{\text{stale}} = f_\theta(o_{t-d}, \ell)
\end{equation}
Note that $h^{\text{stale}}$ is \emph{not detached} from the computation graph; both receive gradients.

\textbf{Vision-Augmented Action Prediction.}
Both sets of planning features are fused with the \emph{same} real-time vision features $z_t = v_\psi(o_t)$:
\begin{equation}
    \hat{a}^{\text{fresh}} = g_\phi(h^{\text{fresh}}, z_t), \quad \hat{a}^{\text{stale}} = g_\phi(h^{\text{stale}}, z_t)
\end{equation}

\textbf{Dual-Path Loss.}
Both predictions are trained toward the same ground-truth action $a_t$:
\begin{equation}
    \mathcal{L} = (1-\lambda)\underbrace{\|\hat{a}^{\text{fresh}} - a_t\|_1}_{\mathcal{L}_{\text{fresh}}} + \lambda\underbrace{\|\hat{a}^{\text{stale}} - a_t\|_1}_{\mathcal{L}_{\text{stale}}}
    \label{eq:loss}
\end{equation}

$\mathcal{L}_{\text{fresh}}$ trains the action head for synchronous operation and provides direct action supervision to the backbone.
$\mathcal{L}_{\text{stale}}$ trains the action head to \emph{compensate} for stale planning features using real-time vision: when $h^{\text{stale}}$ is misaligned with the current state, the action head must rely more heavily on $z_t$.
$\lambda$ balances the two losses; we use curriculum learning to gradually increase $\lambda$ from 0 to $\lambda_{max}$ over the first part of training steps $n_{\text{warmup}}$, allowing the backbone to first learn strong representations before being pressured to be delay-invariant.

\subsection{Deployment Protocol}

\begin{enumerate}
    \item Edge captures observation $o_t$ and sends it to cloud (asynchronous, non-blocking).
    \item When cloud returns $h$ (possibly from a previous $o_{t-k}$): $h_{\text{received}} \leftarrow h$.
    \item Edge computes $z_t = v_\psi(o_t)$ (real-time, local).
    \item Edge computes $\hat{a}_t = g_\phi(h_{\text{received}}, z_t)$.
    \item Edge executes $\hat{a}_t$.
\end{enumerate}
The robot \textbf{never blocks} waiting for the cloud: it always uses the most recently received planning features combined with current real-time vision.


\subsection{Emergent Representational Specialization}
\label{sec:emergent}

The stale path receives an action target from time $t$ but a cloud input from $t-d$:
\begin{equation}
    \nabla_\theta \mathcal{L}_{\mathrm{stale}}
    = \nabla_\theta \|g_\phi(f_\theta(o_{t-d},\ell),v_\psi(o_t))-a_t\|_1.
\end{equation}
Across random $d$, features tied only to the instantaneous state of $o_{t-d}$ are unreliable predictors of $a_t$, whereas task identity, goal, and coarse progress are more stable. The objective therefore \emph{encourages}, but does not mathematically guarantee, invariance to temporal displacement, consistent with the broader connection between nuisance variation and invariant representations~\citep{achille2018emergence}. The current edge feature $z_t$ remains available to encode state-sensitive information.

This design targets complementary robustness at both sides of the cloud-edge interface. Paired-frame training encourages cloud features to be less sensitive to observation age, while the edge-aware action head attenuates residual representation mismatch before it reaches the action output. 
At inference time, the model does not require explicit delay magnitude as input. The edge simply uses whichever cloud feature is most recently available, regardless of how many steps old it is. This is a key distinction from methods that condition policies on measured latency.

\section{Experiments}

\subsection{Implementation}

We build on OpenVLA-OFT~\citep{openvlaoft2025}, which adapts a 7B OpenVLA backbone using LoRA, parallel action-chunk decoding, and continuous L1 regression.
Our modifications include: (1) a new Vision-Augmented Action Head that replaces the original L1RegressionActionHead; (2) a paired-frame dataset pipeline extending the RLDS loader to extract historical frames from the same episode; (3) a dual-path forward pass with dual L1 loss; and (4) a delayed evaluation protocol that simulates network latency.

\subsection{Benchmark}

We evaluate on the \textbf{LIBERO} manipulation benchmark~\citep{liu2024libero}, which consists of 4 task suites (Spatial, Object, Goal, and Long) of 10 tasks each, with 50 demonstration episodes per task.
LIBERO provides a standardized testbed for evaluating language-conditioned manipulation in simulation. 
We compare single-path OpenVLA~\citep{kim2024openvla}, OpenVLA-OFT~\citep{openvlaoft2025}, and UniVLA~\citep{xie2025univla}; future-state-aware asynchronous VLASH~\citep{tang2025vlash}; and CloudEdgeVLA. We feed delayed frames to the single-path policies without changing their inference paths. 

\noindent\textbf{Evaluation Conditions. }
We evaluate under: (1) \textit{No delay ($d_{\max}{=}0$)}, the synchronous reference. (2) \textit{Uniform delay ($k \sim \mathrm{Uniform}\{1,\ldots,d_{\max}\}$)}, with $d_{\max} \in \{5,10,15,20,25,30,40\}$ steps.
We report task success rate (\%) over 50 trials per task. For CloudEdgeVLA, each setting is evaluated with three random seeds, and we report the mean and standard deviation across seeds.

\subsection{Delay Robustness}

\begin{table*}[t]
\centering
\caption{\textbf{LIBERO task success (\%) under uniformly sampled observation delay.}
At $d_{\max}{=}40$, CloudEdgeVLA retains 63.8-78.0\% success across the four suites, whereas VLASH reaches at most 6.4\% and the single-path baselines at most 3.0\%. CloudEdgeVLA entries report mean $\pm$ standard deviation over three seeds; published baseline values are point estimates.}
\label{tab:main_results}
\small
\resizebox{\textwidth}{!}{%
\begin{tabular}{@{}l cccc cccc cccc cccc@{}}
\toprule
& \multicolumn{4}{c}{$d_{\max}{=}0$} & \multicolumn{4}{c}{$d_{\max}{=}10$} & \multicolumn{4}{c}{$d_{\max}{=}20$} & \multicolumn{4}{c}{$d_{\max}{=}40$} \\
\cmidrule(lr){2-5} \cmidrule(lr){6-9} \cmidrule(lr){10-13} \cmidrule(lr){14-17}
\textbf{Method} & Spat. & Obj. & Goal & Long & Spat. & Obj. & Goal & Long & Spat. & Obj. & Goal & Long & Spat. & Obj. & Goal & Long \\
\midrule
OpenVLA     & 84.6 & 71.2 & 77.0 & 56.2 & 6.8  & 1.2  & 2.2  & 5.2  & 0.2  & 0.0  & 0.0  & 0.0  & 0.0  & 0.0  & 0.0  & 0.0  \\
OpenVLA-OFT & \textbf{98.4} & 98.6 & \textbf{97.2} & 93.4 & 10.6 & 3.6  & 26.2 & 8.4  & 0.0  & 0.0  & 4.0  & 1.4  & 0.0  & 0.0  & 0.0  & 0.0  \\
UniVLA      & 96.0 & 96.6 & 94.6 & 93.2 & 27.4 & 34.8 & 48.2 & 35.0 & 0.4  & 2.6  & 21.8 & 2.2  & 0.0  & 0.0  & 3.0  & 0.5  \\
\midrule
VLASH        & 97.3 & \textbf{99.6} & 96.7 & \textbf{93.5} & 60.0 & 62.8 & 56.8 & 45.6 & 5.0 & 8.2 & 28.2 & 9.2 & 0.0 & 0.4 & 6.4 & 0.0 \\
\midrule
\textbf{Ours}                       & 97.9$\pm$0.23 & 97.8$\pm$0.20 & 96.5$\pm$0.31 & 91.7$\pm$0.42 & \textbf{93.6$\pm$0.53} & \textbf{94.4$\pm$0.53} & \textbf{92.9$\pm$0.83} & \textbf{83.2$\pm$0.53} & \textbf{89.6$\pm$1.20} & \textbf{92.1$\pm$0.42} & \textbf{90.9$\pm$0.81} & \textbf{78.1$\pm$0.42} & \textbf{76.4$\pm$0.92} & \textbf{75.6$\pm$0.80} & \textbf{78.0$\pm$1.20} & \textbf{63.8$\pm$1.51} \\
\bottomrule
\end{tabular}
}
\end{table*}

\begin{figure}[!t]
    \centering
    \includegraphics[width=0.5\columnwidth]{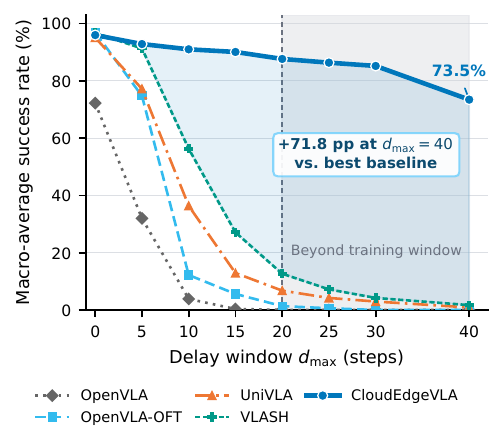}
    \caption{\textbf{Closed-loop uniform-delay sweep on LIBERO.}
    Each point averages Spatial, Object, Goal, and Long. The shaded range $d_{\max}{>}20$ lies beyond training. At $d_{\max}{=}40$, CloudEdgeVLA averages 73.5\% success, 71.8 points above VLASH.}
    \label{fig:closed_loop_delay_curve}
\end{figure}

\begin{figure*}[!t]
    \centering
    \includegraphics[width=\textwidth]{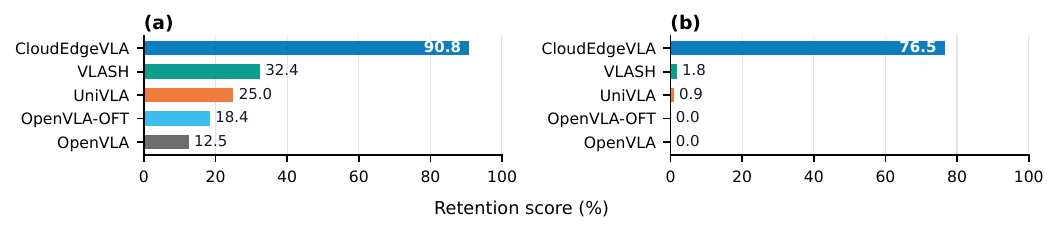}
    \caption{\textbf{Closed-loop delay-retention summaries.}
    Metrics are computed from each method's four-suite macro-average curve. (a) Normalized area under the uniform-delay-window curve. (b) Synchronous success retained at $d_{\max}{=}40$. CloudEdgeVLA achieves 90.8\% and 76.5\%, respectively; VLASH reaches 32.4\% and 1.8\%.}
    \label{fig:closed_loop_retention_summary}
\end{figure*}

Figures~\ref{fig:closed_loop_delay_curve} and~\ref{fig:closed_loop_retention_summary} summarize macro‑averaged outcomes across the four LIBERO suites, while Table~\ref{tab:main_results} presents detailed per‑suite performance. At $d_{\max}{=}10$, OpenVLA achieves at most 6.8\% success, and OpenVLA‑OFT tops out at 26.2\%. Although UniVLA degrades comparatively slowly on the Goal and Long suites, reaching 48.2\% and 35.0\% respectively, its performance collapses to no more than 3.0\% under the heavy $d_{\max}{=}40$ delay setting. VLASH delivers competitive performance under moderate delay, obtaining 45.6–62.8\% success at $d_{\max}{=}10$, yet its success rate plummets to merely 0.0–6.4\% when delay increases to $d_{\max}{=}40$. By contrast, CloudEdgeVLA maintains 76.5\% of its zero‑delay synchronous success rate at $d_{\max}{=}40$, substantially outperforming all other baselines. 

Under \(d_{\text{max}}=40\) uniform delay, CloudEdgeVLA achieves success rates of 76.4\% (Spatial), 75.6\% (Object), 78.0\% (Goal), and 63.8\% (Long). Relative to zero‑delay performance, the success drop is 21.5, 22.2, 18.5, 27.9 percentage points respectively.
The degradation is therefore suite‑dependent: Long shows the largest loss, followed by Object and Spatial, while Goal is the most stable over the tested range. Notably, our model is only trained with a maximum training delay of \(d_{\text{max}}=20\), yet it retains strong task performance under the more challenging test‑time condition \(d_{\text{max}}=40\). Even in the worst case on the Long suite, CloudEdgeVLA remains 63.3 points above the strongest baseline at \(d_{\text{max}}=40\). This demonstrates that our paired‑frame training objective yields meaningful generalization to larger staleness magnitudes not directly seen during training.

\subsection{Ablation Studies}

\begin{table}[t]
\centering
\caption{\textbf{Ablation and real-robot results.}}
\label{tab:combined}
\small
\begin{minipage}{0.48\linewidth}
\centering
\subcaption{Ablation: Vision encoder and loss components on LIBERO-Object under uniform delay $d_{\max}{=}10$.}
\label{tab:ablation}
\setlength{\tabcolsep}{5pt}
\resizebox{0.55\linewidth}{!}{%
\begin{tabular}{@{}l c@{}}
\toprule
\textbf{Variant} & \textbf{Success Rate(\%)} \\
\midrule
No vision encoder     & 80.6 \\
SigLIP-Base          & 94.2 \\
SigLIP-SO400M                 & 94.8 \\
\midrule
$\mathcal{L}_{\text{fresh}}$ only      & 84.6 \\
$\mathcal{L}_{\text{stale}}$ only      & 92.4 \\
$\mathcal{L}_{\text{fresh}} + \mathcal{L}_{\text{stale}}$ & 94.2 \\
\bottomrule
\end{tabular}
}
\end{minipage}
\hfill
\begin{minipage}{0.48\linewidth}
\centering
\subcaption{Real-robot task success (\%). Cells report static/dynamic success over 10 trials per variant. CE-VLA denotes CloudEdgeVLA. Round-trip time (RTT) is the emulated delay added to the native serving pipeline.}
\label{tab:real_robot}
\setlength{\tabcolsep}{5pt}
\resizebox{0.55\linewidth}{!}{%
\begin{tabular}{@{}lccc@{}}
\toprule
& \multicolumn{3}{c}{\textbf{Added network RTT (ms)}} \\
\cmidrule(lr){2-4}
\textbf{Method} & \textbf{0} & \textbf{400} & \textbf{1000}\\
\midrule
VLASH & 100/90 & 60/30 & 0/0 \\
\midrule
\textbf{CE-VLA} & \textbf{100/90} & \textbf{90/90} & \textbf{80/70}\\
\bottomrule
\end{tabular}
}
\end{minipage}
\end{table}

We isolate the contribution of current edge vision, encoder capacity, and the two loss paths on the LIBERO-Object suite. Table~\ref{tab:ablation} reports success under uniform delay with $d_{\max}{=}10$; all variants use the same backbone, demonstrations, and optimization budget. 
The decisive architectural factor is whether the action head receives a current observation. Adding the frozen SigLIP-Base encoder used by our default model raises success from 80.6\% to 94.2\%, a 13.6-point gain over stale cloud features alone. Replacing it with the substantially larger SigLIP-SO400M reaches 94.8\%, only 0.6 points higher. Thus, most of the benefit comes from real-time visual grounding rather than encoder scale, supporting the lightweight Base model for edge deployment. The loss ablation shows a complementary pattern under this default configuration. Stale-only supervision reaches 92.4\%, 7.8 points above fresh-only training, confirming that explicit exposure to delayed features drives most of the robustness. Joint supervision restores the full model to 94.2\%, another 1.8-point gain over stale-only training; the fresh path therefore provides complementary grounding rather than replacing delay exposure.

\subsection{Real-Robot Sanity Check}

We conduct a small Franka pick-and-place pilot to check physical feasibility. An RTX~5080 workstation runs the edge vision encoder, action head, and control loop, while the 7B backbone is served from an RTX~4090. The robot must place a toy bear into a box. We test a static setting and a dynamic variant in which the task target is displaced within 10\,cm during reaching. A rollout succeeds when the bear is released inside the box. Each method receives 10 trials per variant under added RTT of 0, 400, and 1000\,ms. These values are emulated delays added on top of the native system latency: 0\,ms means no \emph{additional} RTT, not zero camera-to-action latency. Camera acquisition, preprocessing, transport, request scheduling, and model inference remain present in every profile.

\begin{figure*}[t]
    \centering
    \includegraphics[width=0.98\textwidth]{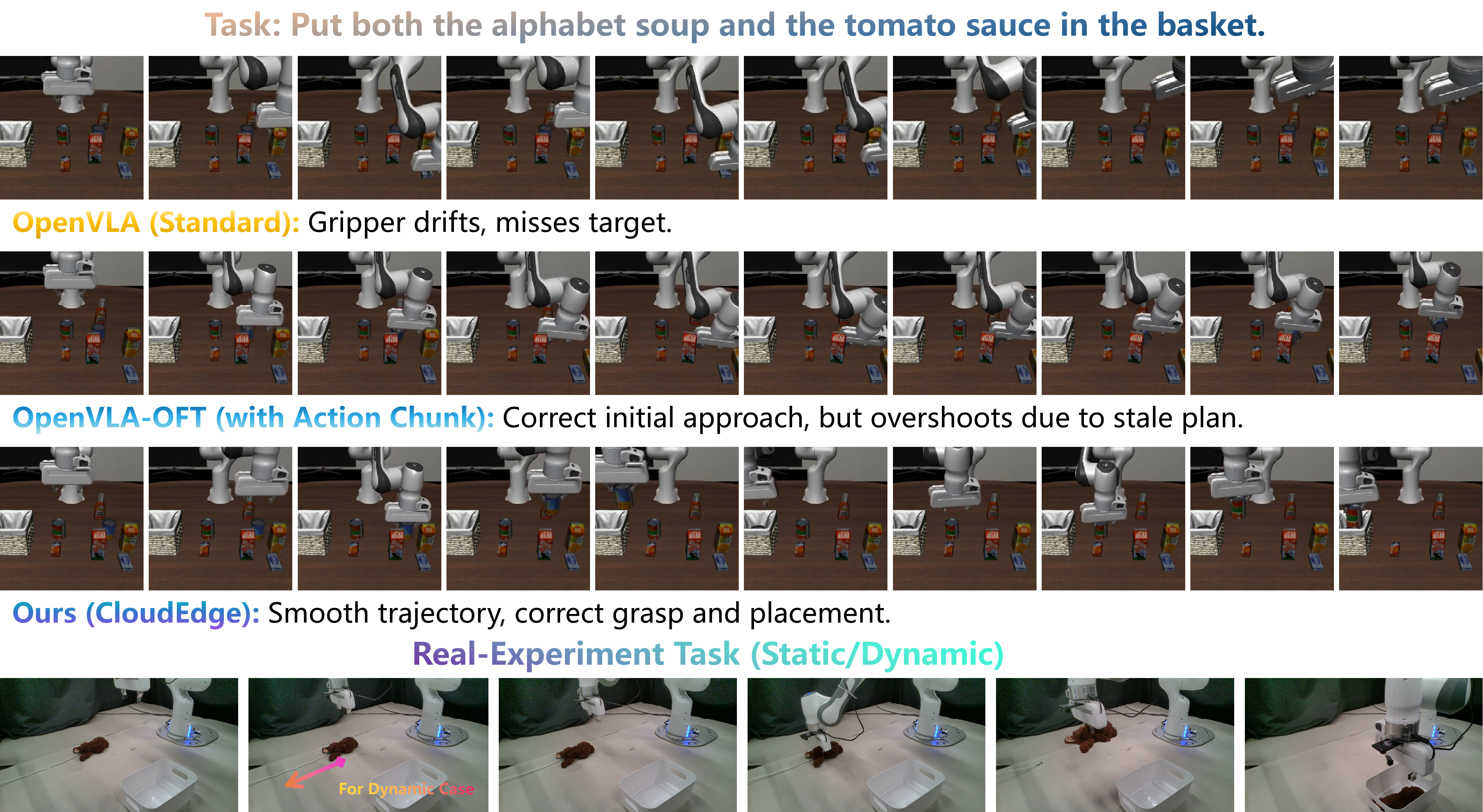}
    \caption{\textbf{Qualitative rollouts in simulation and on the real robot.}
    The top three sequences compare OpenVLA, OpenVLA-OFT, and CloudEdgeVLA on the same LIBERO task of placing two specified objects in the basket. The bottom sequence shows CloudEdgeVLA executing the real-robot task of placing a toy bear into a box; the arrows mark the externally displaced target in the dynamic setting.}
    \label{fig:qualitative}
\end{figure*}

The results are summarized in Table~\ref{tab:real_robot}. With zero added RTT, CE‑VLA and VLASH achieve 100\%/90\% success on static and dynamic tasks. At 400\,ms emulated RTT, CE‑VLA maintains 90\%/90\% success, outperforming VLASH (60\%/30\%) by 30 and 60 percentage points. At 1000\,ms RTT, CE‑VLA retains 80\%/70\% success, while VLASH degrades completely to 0\%/0\%. This widening performance gap under elevated latency further validates that our cloud‑edge design preserves manipulation performance even under severe network round‑trip delays.
Figure~\ref{fig:qualitative} complements these success rates with representative simulation and real-robot rollouts.

\section{Representation Analysis: Backbone‑Head Delay Mechanism}
\label{sec:repr_analysis}

To verify the effectiveness of our delay‑robust feature learning and uncover the underlying robustness mechanism, we perform controlled representation analysis comparing the original OpenVLA‑OFT checkpoint against our CloudEdgeVLA checkpoint. 
We evaluate over 80 shared demonstration timesteps sampled across the ten LIBERO‑Spatial tasks. For each timestep, we feed the cloud backbone a frame delayed by $d$ steps while keeping proprioceptive inputs unchanged.
We quantify backbone representation drift and end‑to‑end action drift via
\begin{equation}
\begin{split}
    D_h(d) &= \mathbb{E}\left[1-\cos\!\left(h_t,h_{t-d}\right)\right], \\
    D_a(d) &= \mathbb{E}\left[\left|\hat{a}_t^{(0)}-\hat{a}_t^{(d)}\right|\right],
\end{split}
\end{equation}
and further define the staleness transfer gain
\begin{equation}
    \kappa(d)=\frac{D_a(d)}{D_h(d)+\epsilon}.
\end{equation}



A lower $D_h$ indicates the backbone yields temporally stable features under delayed observations, while a lower $\kappa$ shows the edge head constrains residual backbone drift from propagating into action outputs. The supplementary material plots $D_a$ and normalized MAE versus demonstrated action chunks, ruling out the trivial hypothesis that gains stem from an overly insensitive policy emitting near‑constant actions.

\begin{figure}[!t]
    \centering
    \begin{minipage}[t]{0.48\linewidth}
        \centering
        \includegraphics[width=\linewidth]{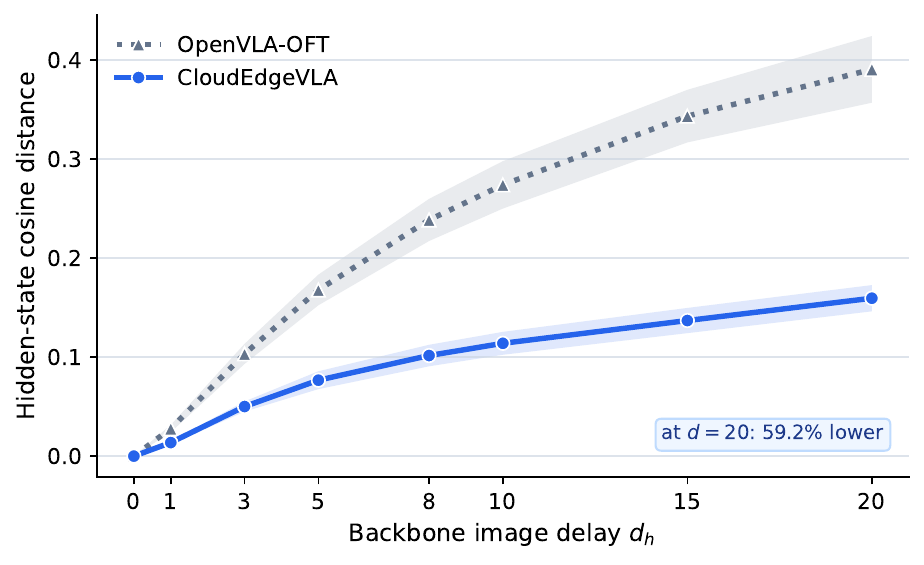}
        \captionsetup{font=footnotesize}
        \captionof{figure}{\textbf{Backbone representation staleness on LIBERO-Spatial.}
        Over 80 shared states from ten tasks, CloudEdgeVLA reduces backbone drift $D_h$ by 59.2\% at $d{=}20$; bands are task-level 95\% confidence intervals.}
        \label{fig:backbone_staleness}
    \end{minipage}\hfill
    \begin{minipage}[t]{0.48\linewidth}
        \centering
        \includegraphics[width=\linewidth]{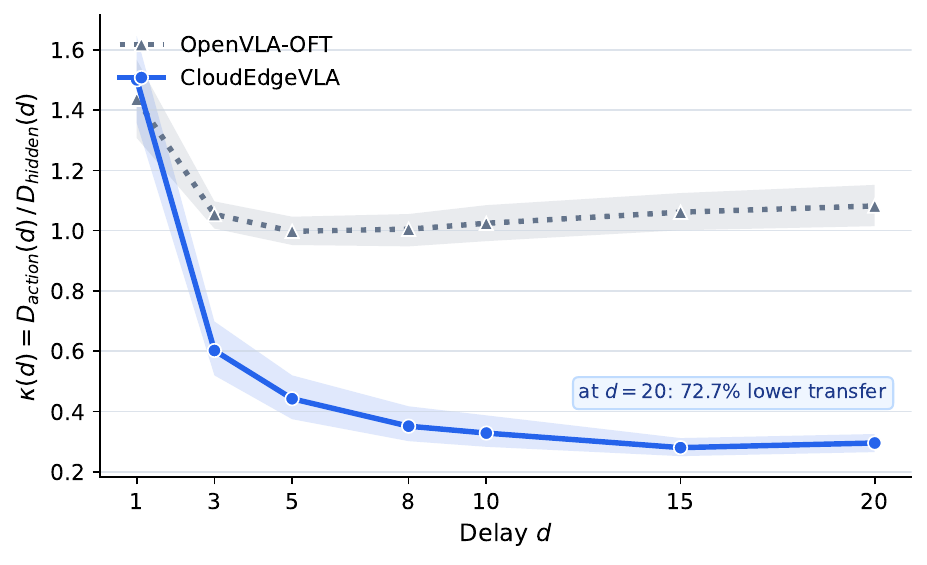}
        \captionsetup{font=footnotesize}
        \captionof{figure}{\textbf{Head staleness-transfer gain on LIBERO-Spatial.}
        CloudEdgeVLA reduces $\kappa$, the action drift transferred per unit backbone drift, by 72.7\% at $d{=}20$.}
        \label{fig:head_staleness_transfer}
    \end{minipage}
\end{figure}

Figures~\ref{fig:backbone_staleness} and~\ref{fig:head_staleness_transfer} decompose robustness sources across the cloud‑edge interface. At $d{=}20$, CloudEdgeVLA reduces backbone drift $D_h$ from 0.391 to 0.160 and lowers staleness‑transfer gain $\kappa$ from 1.082 to 0.295. Together these two factors suppress end‑to‑end action drift $D_a$ from 0.423 to 0.047, with demonstration MAE falling from 0.421 to 0.048. The findings validate two complementary properties of our learned interface: the cloud backbone learns delay‑robust task representations, and the lightweight edge head mitigates leftover drift within received cloud features. The supplement further visualizes these curves, the small accuracy penalty on fresh inputs, per‑task consistency, and action‑chunk effects.

\section{Discussion and Limitations}
CloudEdgeVLA re‑designs the slow‑fast cloud‑edge interface. Rather than sending complete temporally aligned action plans from cloud to edge, we transmit delay‑resilient task‑level representations. The edge then grounds these representations using real‑time local visual observations.
The state‑sensitive decision‑making resides on local edge hardware. Unlike action‑chunk open‑loop execution, our setup maintains visual feedback closure on the edge. 
This yields a compact, auditable edge stack, while cloud‑hosted foundation models are free to scale independently without imposing heavier computation requirements on the robot side.

The evidence has three boundaries. First, robustness is established only for the tested delay distributions; an extended disconnection can invalidate even task-level context. 
Second, paired‑frame training uses two backbone passes per training sample, which increases training compute overhead.
Third, a frozen RGB encoder may miss depth, force, or contact cues that would benefit dynamic manipulation.
Timestamped features and a local fallback policy remain necessary for handling prolonged network disconnection in real deployments.

\section{Conclusion}


CloudEdgeVLA enables a cloud‑hosted VLA and lightweight edge controller to operate without blocking synchronization. Our paired‑frame training objective makes stale cloud‑derived features and current local vision jointly predictive of the current robot action. 
This design retains high task success under a 40‑step uniform‑delay window, twice the maximum staleness encountered during training. It demonstrates generalization to staleness magnitudes outside the training distribution and substantially outperforms all VLA baselines.
A small‑scale real‑robot pilot further provides a physical feasibility check. Collectively, our findings frame latency tolerance for cloud‑edge robot control as a learned property, rather than merely an inference‑scheduling problem.

\FloatBarrier
\bibliographystyle{iclr2027_conference}
\bibliography{iclr2027_conference}

@inproceedings{brohan2023rt2,
  title     = {{RT-2}: Vision-Language-Action Models Transfer Web Knowledge to Robotic Control},
  author    = {Zitkovich, Brianna and Yu, Tianhe and Xu, Sichun and Xu, Peng and Xiao, Ted and Xia, Fei and Wu, Jialin and Wohlhart, Paul and Welker, Stefan and Wahid, Ayzaan and others},
  booktitle = {Proceedings of the 7th Conference on Robot Learning},
  series    = {Proceedings of Machine Learning Research},
  volume    = {229},
  pages     = {2165--2183},
  year      = {2023},
  url       = {https://proceedings.mlr.press/v229/zitkovich23a.html}
}

@inproceedings{kim2024openvla,
  title     = {{OpenVLA}: An Open-Source Vision-Language-Action Model},
  author    = {Kim, Moo Jin and Pertsch, Karl and Karamcheti, Siddharth and Xiao, Ted and Balakrishna, Ashwin and Nair, Suraj and Rafailov, Rafael and Foster, Ethan P. and Sanketi, Pannag R. and Vuong, Quan and others},
  booktitle = {Proceedings of the 8th Conference on Robot Learning},
  series    = {Proceedings of Machine Learning Research},
  volume    = {270},
  pages     = {2679--2713},
  year      = {2025},
  url       = {https://proceedings.mlr.press/v270/kim25c.html}
}

@article{black2024pi0,
  title   = {$\pi_0$: A Vision-Language-Action Flow Model for General Robot Control},
  author  = {Black, Kevin and Brown, Noah and Driess, Danny and Esmail, Adnan and Equi, Michael and Finn, Chelsea and Fusai, Niccolo and Groom, Lachy and Hausman, Karol and Ichter, Brian and others},
  journal = {arXiv preprint arXiv:2410.24164},
  year    = {2024},
  url     = {https://arxiv.org/abs/2410.24164}
}

@article{openvlaoft2025,
  title   = {Fine-Tuning Vision-Language-Action Models: Optimizing Speed and Success},
  author  = {Kim, Moo Jin and Finn, Chelsea and Liang, Percy},
  journal = {arXiv preprint arXiv:2502.19645},
  year    = {2025},
  url     = {https://arxiv.org/abs/2502.19645}
}

@inproceedings{team2024octo,
  title     = {{Octo}: An Open-Source Generalist Robot Policy},
  author    = {Ghosh, Dibya and Walke, Homer Rich and Pertsch, Karl and Black, Kevin and Mees, Oier and Dasari, Sudeep and Hejna, Joey and Kreiman, Tobias and Xu, Charles and Luo, Jianlan and others},
  booktitle = {Proceedings of Robotics: Science and Systems},
  year      = {2024},
  doi       = {10.15607/RSS.2024.XX.090}
}

@article{xie2025univla,
  title   = {Unified Vision-Language-Action Model},
  author  = {Wang, Yuqi and Li, Xinghang and Wang, Wenxuan and Zhang, Junbo and Li, Yingyan and Chen, Yuntao and Wang, Xinlong and Zhang, Zhaoxiang},
  journal = {arXiv preprint arXiv:2506.19850},
  year    = {2025},
  url     = {https://arxiv.org/abs/2506.19850}
}

@inproceedings{chi2023diffusionpolicy,
  title     = {Diffusion Policy: Visuomotor Policy Learning via Action Diffusion},
  author    = {Chi, Cheng and Feng, Siyuan and Du, Yilun and Xu, Zhenjia and Cousineau, Eric and Burchfiel, Benjamin C. M. and Song, Shuran},
  booktitle = {Proceedings of Robotics: Science and Systems},
  year      = {2023},
  doi       = {10.15607/RSS.2023.XIX.026}
}

@inproceedings{karkus2024actionchunk,
  title     = {Learning Fine-Grained Bimanual Manipulation with Low-Cost Hardware},
  author    = {Zhao, Tony Z. and Kumar, Vikash and Levine, Sergey and Finn, Chelsea},
  booktitle = {Proceedings of Robotics: Science and Systems},
  year      = {2023},
  doi       = {10.15607/RSS.2023.XIX.016}
}

@inproceedings{saxena2023mrest,
  title     = {Multi-Resolution Sensing for Real-Time Control with Vision-Language Models},
  author    = {Saxena, Saumya and Sharma, Mohit and Kroemer, Oliver},
  booktitle = {Proceedings of the 7th Conference on Robot Learning},
  series    = {Proceedings of Machine Learning Research},
  volume    = {229},
  pages     = {2210--2228},
  year      = {2023},
  url       = {https://proceedings.mlr.press/v229/saxena23a.html}
}

@article{han2024dpvla,
  title   = {A Dual Process {VLA}: Efficient Robotic Manipulation Leveraging {VLM}},
  author  = {Han, ByungOk and Kim, Jaehong and Jang, Jinhyeok},
  journal = {arXiv preprint arXiv:2410.15549},
  year    = {2024},
  url     = {https://arxiv.org/abs/2410.15549}
}

@inproceedings{zhang2025hirt,
  title     = {{HiRT}: Enhancing Robotic Control with Hierarchical Robot Transformers},
  author    = {Zhang, Jianke and Guo, Yanjiang and Chen, Xiaoyu and Wang, Yen-Jen and Hu, Yucheng and Shi, Chengming and Chen, Jianyu},
  booktitle = {Proceedings of the 8th Conference on Robot Learning},
  series    = {Proceedings of Machine Learning Research},
  volume    = {270},
  pages     = {933--946},
  year      = {2025},
  url       = {https://proceedings.mlr.press/v270/zhang25b.html}
}

@article{shukor2025smolvla,
  title   = {{SmolVLA}: A Vision-Language-Action Model for Affordable and Efficient Robotics},
  author  = {Shukor, Mustafa and Aubakirova, Dana and Capuano, Francesco and Kooijmans, Pepijn and Palma, Steven and Zouitine, Adil and Aractingi, Michel and Pascal, Caroline and Russi, Martino and Marafioti, Andres and others},
  journal = {arXiv preprint arXiv:2506.01844},
  year    = {2025},
  url     = {https://arxiv.org/abs/2506.01844}
}

@article{chen2025fastinslow,
  title   = {Fast-in-Slow: A Dual-System Foundation Model Unifying Fast Manipulation within Slow Reasoning},
  author  = {Chen, Hao and Liu, Jiaming and Gu, Chenyang and Liu, Zhuoyang and Zhang, Renrui and Li, Xiaoqi and He, Xiao and Guo, Yandong and Fu, Chi-Wing and Zhang, Shanghang and Heng, Pheng-Ann},
  journal = {arXiv preprint arXiv:2506.01953},
  year    = {2025},
  url     = {https://arxiv.org/abs/2506.01953}
}

@article{sendai2025a2c2,
  title   = {Leave No Observation Behind: Real-Time Correction for {VLA} Action Chunks},
  author  = {Sendai, Kohei and Alvarez, Maxime and Matsushima, Tatsuya and Matsuo, Yutaka and Iwasawa, Yusuke},
  journal = {arXiv preprint arXiv:2509.23224},
  year    = {2025},
  url     = {https://arxiv.org/abs/2509.23224}
}

@article{tang2025vlash,
  title   = {{VLASH}: Real-Time {VLA}s via Future-State-Aware Asynchronous Inference},
  author  = {Tang, Jiaming and Sun, Yufei and Zhao, Yilong and Yang, Shang and Lin, Yujun and Zhang, Zhuoyang and Hou, James and Lu, Yao and Liu, Zhijian and Han, Song},
  journal = {arXiv preprint arXiv:2512.01031},
  year    = {2025},
  url     = {https://arxiv.org/abs/2512.01031}
}

@article{yan2026acting,
  title   = {Acting While Understanding: Asynchronous Semantic-Action Decoupling for Real-Time Vision-Language-Action Models},
  author  = {Yan, Shenhao and Wang, Ge and Liu, Qi and Meng, Weilin and Yang, Jiahao and Yao, Chengsi and Feng, Fan and Ma, Xiaoguang and Zhao, Yiming and Han, Yatong},
  journal = {arXiv preprint arXiv:2606.15285},
  year    = {2026},
  url     = {https://arxiv.org/abs/2606.15285}
}

@inproceedings{xiao2020concurrent,
  title     = {Thinking While Moving: Deep Reinforcement Learning with Concurrent Control},
  author    = {Xiao, Ted and Jang, Eric and Kalashnikov, Dmitry and Levine, Sergey and Ibarz, Julian and Hausman, Karol and Herzog, Alexander},
  booktitle = {International Conference on Learning Representations},
  year      = {2020},
  url       = {https://openreview.net/forum?id=SJexHkSFPS}
}

@inproceedings{wang2024signal,
  title     = {Addressing Signal Delay in Deep Reinforcement Learning},
  author    = {Wang, Wei and Han, Dongqi and Luo, Xufang and Li, Dongsheng},
  booktitle = {International Conference on Learning Representations},
  year      = {2024},
  url       = {https://openreview.net/forum?id=Z8UfDs4J46}
}

@inproceedings{karamzade2024worldmodels,
  title     = {Reinforcement Learning from Delayed Observations via World Models},
  author    = {Karamzade, Armin and Kim, Kyungmin and Kalsi, Montek and Fox, Roy},
  booktitle = {Reinforcement Learning Conference},
  year      = {2024},
  url       = {https://openreview.net/forum?id=ToftWbZfYr}
}

@article{kehoe2015cloudrobotics,
  title   = {A Survey of Research on Cloud Robotics and Automation},
  author  = {Kehoe, Ben and Patil, Sachin and Abbeel, Pieter and Goldberg, Ken},
  journal = {IEEE Transactions on Automation Science and Engineering},
  volume  = {12},
  number  = {2},
  pages   = {398--409},
  year    = {2015},
  doi     = {10.1109/TASE.2014.2376492}
}

@article{wan2016cloudrobotics,
  title   = {Cloud Robotics: Current Status and Open Issues},
  author  = {Wan, Jiafu and Tang, Shenglong and Yan, Hehua and Li, Di and Wang, Shiyong and Vasilakos, Athanasios V.},
  journal = {IEEE Access},
  volume  = {4},
  pages   = {2797--2807},
  year    = {2016},
  doi     = {10.1109/ACCESS.2016.2574979}
}

@inproceedings{chinchali2019offloading,
  title     = {Network Offloading Policies for Cloud Robotics: A Learning-Based Approach},
  author    = {Chinchali, Sandeep and Sharma, Apoorva and Harrison, James and Elhafsi, Amine and Kang, Daniel and Pergament, Evgenya and Cidon, Eyal and Katti, Sachin and Pavone, Marco},
  booktitle = {Proceedings of Robotics: Science and Systems},
  year      = {2019},
  doi       = {10.15607/RSS.2019.XV.063}
}

@inproceedings{liu2024libero,
  title     = {{LIBERO}: Benchmarking Knowledge Transfer for Lifelong Robot Learning},
  author    = {Liu, Bo and Zhu, Yifeng and Gao, Chongkai and Feng, Yihao and Liu, Qiang and Zhu, Yuke and Stone, Peter},
  booktitle = {Advances in Neural Information Processing Systems},
  volume    = {36},
  year      = {2023},
  url       = {https://proceedings.neurips.cc/paper_files/paper/2023/hash/8c3c666820ea055a77726d66fc7d447f-Abstract-Datasets_and_Benchmarks.html}
}

@inproceedings{zhai2023siglip,
  title     = {Sigmoid Loss for Language Image Pre-Training},
  author    = {Zhai, Xiaohua and Mustafa, Basil and Kolesnikov, Alexander and Beyer, Lucas},
  booktitle = {Proceedings of the IEEE/CVF International Conference on Computer Vision},
  pages     = {11975--11986},
  year      = {2023},
  url       = {https://openaccess.thecvf.com/content/ICCV2023/html/Zhai_Sigmoid_Loss_for_Language_Image_Pre-Training_ICCV_2023_paper.html}
}

@inproceedings{hu2022lora,
  title     = {{LoRA}: Low-Rank Adaptation of Large Language Models},
  author    = {Hu, Edward J. and Shen, Yelong and Wallis, Phillip and Allen-Zhu, Zeyuan and Li, Yuanzhi and Wang, Shean and Wang, Lu and Chen, Weizhu},
  booktitle = {International Conference on Learning Representations},
  year      = {2022},
  url       = {https://openreview.net/forum?id=nZeVKeeFYf9}
}

@article{achille2018emergence,
  title   = {Emergence of Invariance and Disentanglement in Deep Representations},
  author  = {Achille, Alessandro and Soatto, Stefano},
  journal = {Journal of Machine Learning Research},
  volume  = {19},
  number  = {50},
  pages   = {1--34},
  year    = {2018},
  url     = {https://jmlr.org/papers/v19/17-646.html}
}

@book{kahneman2011thinking,
  title     = {Thinking, Fast and Slow},
  author    = {Kahneman, Daniel},
  publisher = {Farrar, Straus and Giroux},
  year      = {2011}
}

\ifdefined\CloudEdgeIncludeAppendix
\FloatBarrier
\ifdefined\CloudEdgeAppendixPageBreak
\clearpage
\fi
\appendix
\setcounter{secnumdepth}{1}
\def\CloudEdgeAppendixBodyOnly{}
\ifdefined\CloudEdgeAppendixBodyOnly
\else
\documentclass{article}
\usepackage{iclr2027_conference,times}
\usepackage[hyphens]{url}
\usepackage{hyperref}
\usepackage{graphicx}
\usepackage{caption}
\usepackage{placeins}
\usepackage{float}
\usepackage{amsmath}
\usepackage{amssymb}
\usepackage{booktabs}
\usepackage{xcolor}
\urlstyle{rm}
\def\UrlFont{\rm}
\frenchspacing
\raggedbottom

\title{Supplementary Material: Delay-Mechanism Analysis of CloudEdgeVLA}

\begin{document}
\maketitle
\fi

\section{Offline Mechanism-Diagnostic Protocol}

This supplement expands the backbone--head analysis in the main paper using the aligned CloudEdgeVLA checkpoint at 120k training steps and the original OpenVLA-OFT checkpoint. We use one demonstration episode from each of the ten LIBERO-Spatial tasks and select eight valid current timesteps per task, giving 80 shared evaluation states. For this offline diagnostic, we replace the cloud image by a frame at each deterministic offset
\[
    d \in \{0,1,3,5,8,10,15,20\}
\]
environment steps. These fixed offsets resolve how feature and action drift develop within the $W{=}21$ paired-frame training window; they are mechanism probes, not additional closed-loop delay conditions. The closed-loop evaluation in the main paper instead samples observation age as $k\sim\mathrm{Uniform}\{1,\ldots,d_{\max}\}$. Proprioception remains current for both models, and CloudEdgeVLA additionally receives the current image through its edge encoder. Predictions are normalized eight-step action chunks with seven action dimensions.

Let $h_t$ denote the action-token hidden states produced from the current image and $h_{t-d}$ those produced from the delayed image. We report
\begin{align}
    D_h(d) &= \mathbb{E}\left[1-\cos\!\left(h_t,h_{t-d}\right)\right], \\
    D_a(d) &= \mathbb{E}\left[\left|\hat{a}_t^{(0)}-\hat{a}_t^{(d)}\right|\right], \\
    \kappa(d) &= \frac{D_a(d)}{D_h(d)+\epsilon}, \\
    E_{\mathrm{demo}}(d) &= \mathbb{E}\left[\left|\hat{a}_t^{(d)}-a_t\right|\right].
\end{align}
$D_h$ measures backbone representation staleness, $\kappa$ measures how strongly the head transfers residual hidden-state drift into the action output, and $E_{\mathrm{demo}}$ prevents a nearly constant predictor from appearing robust merely because its output changes little. Confidence intervals in the figures are computed across the ten task means.

\begin{table}[H]
    \centering
    \caption{\textbf{Fixed-offset offline diagnostics on LIBERO-Spatial.} OFT denotes original OpenVLA-OFT and CE denotes aligned CloudEdgeVLA at 120k steps. All entries are means over 80 shared demonstration states. The transfer gain is undefined at $d{=}0$ because both drift terms are zero.}
    \label{tab:supp_delay_metrics}
    {\scriptsize
    \setlength{\tabcolsep}{4.5pt}
    \renewcommand{\arraystretch}{0.95}
    \begin{tabular}{@{}r cc cc cc cc@{}}
        \toprule
        & \multicolumn{2}{c}{$D_h$} & \multicolumn{2}{c}{$\kappa$} & \multicolumn{2}{c}{$D_a$} & \multicolumn{2}{c}{$E_{\mathrm{demo}}$} \\
        \cmidrule(lr){2-3}\cmidrule(lr){4-5}\cmidrule(lr){6-7}\cmidrule(lr){8-9}
        $d$ & OFT & CE & OFT & CE & OFT & CE & OFT & CE \\
        \midrule
        0  & 0.000 & 0.000 & --    & --    & 0.000 & 0.000 & 0.020 & 0.023 \\
        1  & 0.028 & 0.014 & 1.436 & 1.500 & 0.040 & 0.021 & 0.041 & 0.028 \\
        3  & 0.103 & 0.050 & 1.054 & 0.602 & 0.109 & 0.030 & 0.107 & 0.035 \\
        5  & 0.168 & 0.077 & 0.998 & 0.442 & 0.167 & 0.034 & 0.165 & 0.040 \\
        8  & 0.239 & 0.102 & 1.005 & 0.351 & 0.240 & 0.036 & 0.238 & 0.041 \\
        10 & 0.274 & 0.114 & 1.024 & 0.328 & 0.281 & 0.037 & 0.279 & 0.042 \\
        15 & 0.343 & 0.137 & 1.061 & 0.280 & 0.365 & 0.038 & 0.363 & 0.041 \\
        20 & 0.391 & 0.160 & 1.082 & 0.295 & 0.423 & 0.047 & 0.421 & 0.048 \\
        \bottomrule
    \end{tabular}
    }
\end{table}

Table~\ref{tab:supp_delay_metrics} gives the numerical counterpart to the main paper's mechanism figure. At $d{=}20$, CloudEdgeVLA reduces $D_h$, $\kappa$, $D_a$, and $E_{\mathrm{demo}}$ by 59.2\%, 72.7\%, 88.9\%, and 88.5\%, respectively.

\section{Offline Action-Output Robustness}

\begin{figure}[H]
    \centering
    \begin{minipage}[t]{0.47\textwidth}
        \centering
        \includegraphics[width=\linewidth]{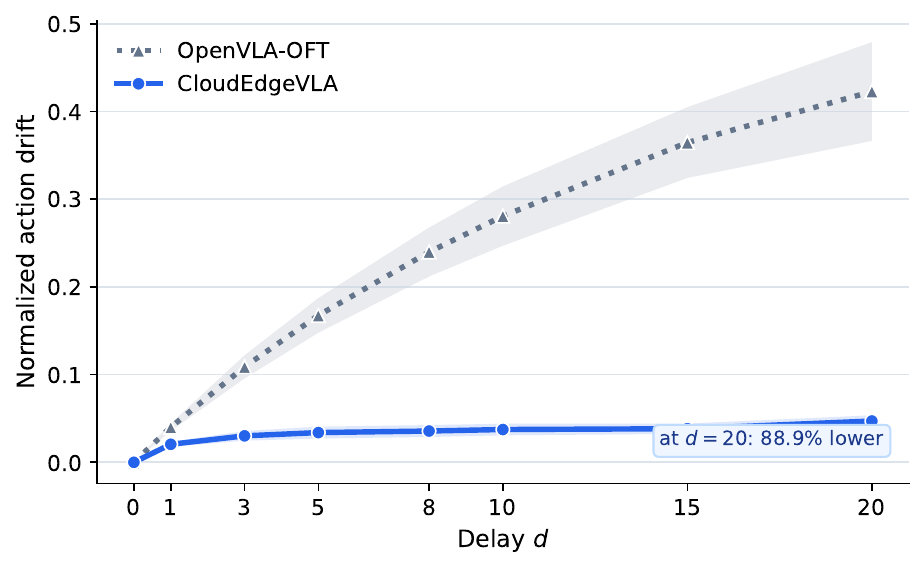}
        \captionsetup{font=footnotesize}
        \captionof{figure}{\textbf{Action drift under delayed vision.}
        Bands are task-level 95\% confidence intervals. At $d{=}20$, CloudEdgeVLA reduces drift from its fresh prediction by 88.9\%.}
        \label{fig:supp_action_drift}
    \end{minipage}\hfill
    \begin{minipage}[t]{0.47\textwidth}
        \centering
        \includegraphics[width=\linewidth]{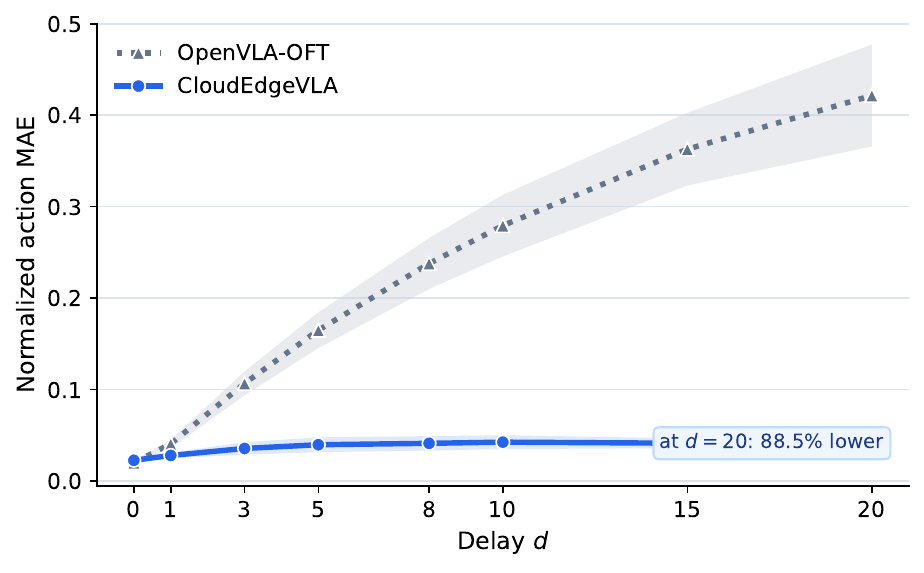}
        \captionsetup{font=footnotesize}
        \captionof{figure}{\textbf{Normalized demonstration MAE under delayed vision.}
        CloudEdgeVLA incurs a small fresh-accuracy cost but becomes better by $d{=}1$ and reduces MAE by 88.5\% at $d{=}20$.}
        \label{fig:supp_demo_mae}
    \end{minipage}
\end{figure}

Figures~\ref{fig:supp_action_drift} and~\ref{fig:supp_demo_mae} expose the tradeoff hidden by a drift-only metric. At $d{=}0$, demonstration MAE is 0.023 for CloudEdgeVLA and 0.020 for OpenVLA-OFT; CloudEdgeVLA becomes better by $d{=}1$, and the gap then widens over the tested range. 
This shows our delay‑aware training incurs only a modest drop in fresh‑observation accuracy while yielding drastically improved robustness, ruling out the trivial explanation of an insensitive, near‑constant action head.

\FloatBarrier

\section{Per-Task Feature-to-Action Geometry}

\begin{figure}[H]
    \centering
    \includegraphics[width=0.82\columnwidth]{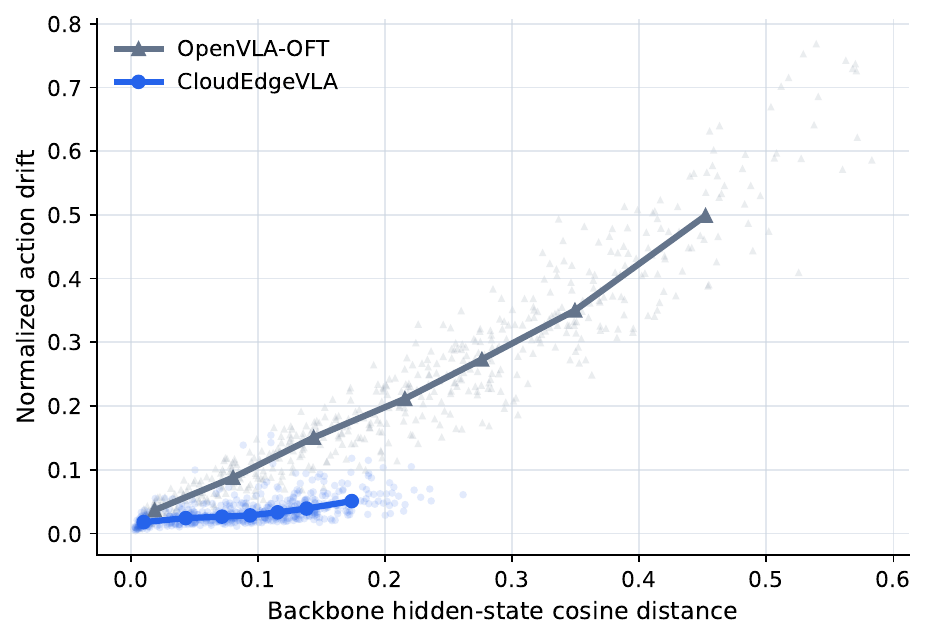}
    \caption{\textbf{Feature-to-action delay geometry.}
    Action drift grows with hidden-state distance for OpenVLA-OFT but remains comparatively flat for CloudEdgeVLA.}
    \label{fig:supp_geometry_attenuation}
\end{figure}

\begin{figure}[H]
    \centering
    \begin{minipage}[t]{0.47\textwidth}
        \centering
        \includegraphics[width=\linewidth]{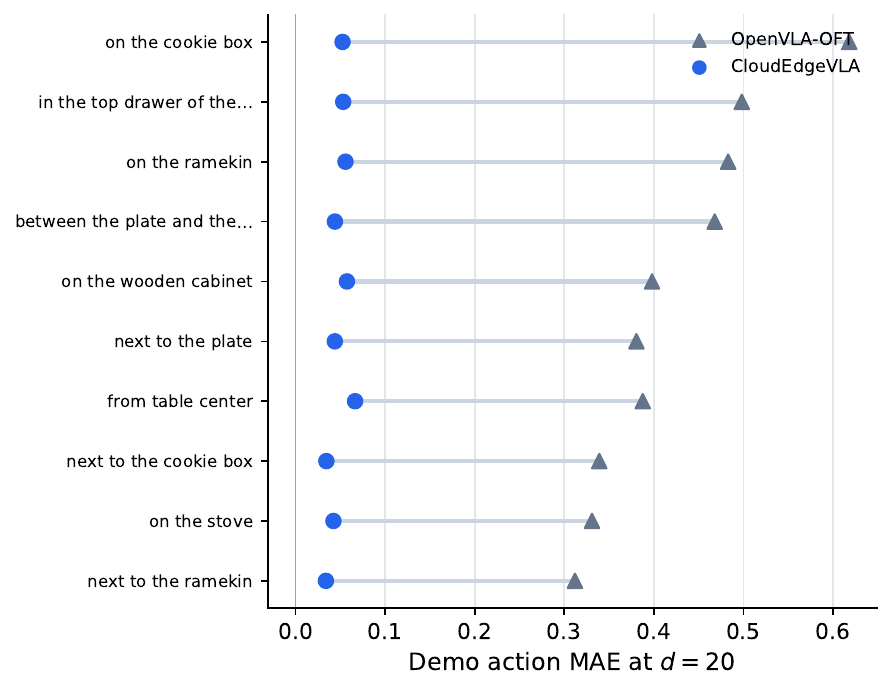}
        \captionsetup{font=footnotesize}
        \captionof{figure}{\textbf{Per-task delayed action error at $d{=}20$.}
        CloudEdgeVLA has lower demonstration MAE on every LIBERO-Spatial task.}
        \label{fig:supp_geometry_task_error}
    \end{minipage}\hfill
    \begin{minipage}[t]{0.47\textwidth}
        \centering
        \includegraphics[width=\linewidth]{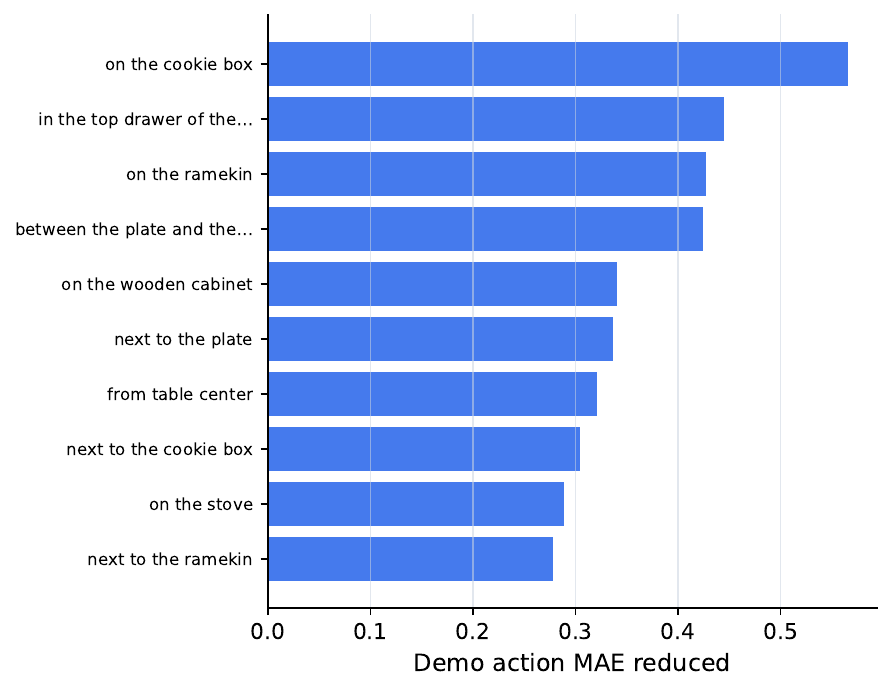}
        \captionsetup{font=footnotesize}
        \captionof{figure}{\textbf{Per-task robustness gain at $d{=}20$.}
        Positive bars denote lower demonstration MAE for CloudEdgeVLA.}
        \label{fig:supp_geometry_gain}
    \end{minipage}
\end{figure}

Figures~\ref{fig:supp_geometry_attenuation}--\ref{fig:supp_geometry_gain} show that CloudEdgeVLA's lower aggregate error is not driven by a small subset of tasks. At $d{=}20$, it has lower demonstration action MAE on all ten tasks. The absolute task-level reduction ranges from 0.278 to 0.566 normalized MAE. The feature-to-action plot also reveals a qualitative difference in propagation: OpenVLA-OFT action drift grows approximately with its hidden-state distance, whereas CloudEdgeVLA's action drift remains comparatively flat as its backbone representation ages.

\section{Action-Chunk Decomposition}

\begin{figure}[H]
    \centering
    \begin{minipage}[t]{0.47\textwidth}
        \centering
        \includegraphics[width=\linewidth]{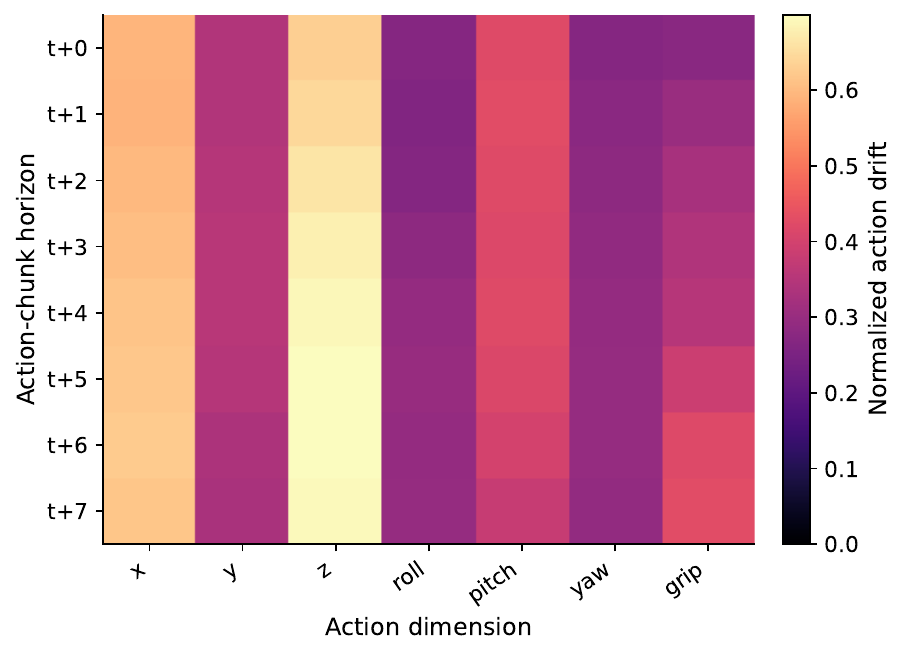}
        \captionsetup{font=footnotesize}
        \captionof{figure}{\textbf{OpenVLA-OFT action-chunk drift at $d{=}20$.}
        Cells show mean absolute drift from the fresh prediction.}
        \label{fig:supp_chunk_oft}
    \end{minipage}\hfill
    \begin{minipage}[t]{0.47\textwidth}
        \centering
        \includegraphics[width=\linewidth]{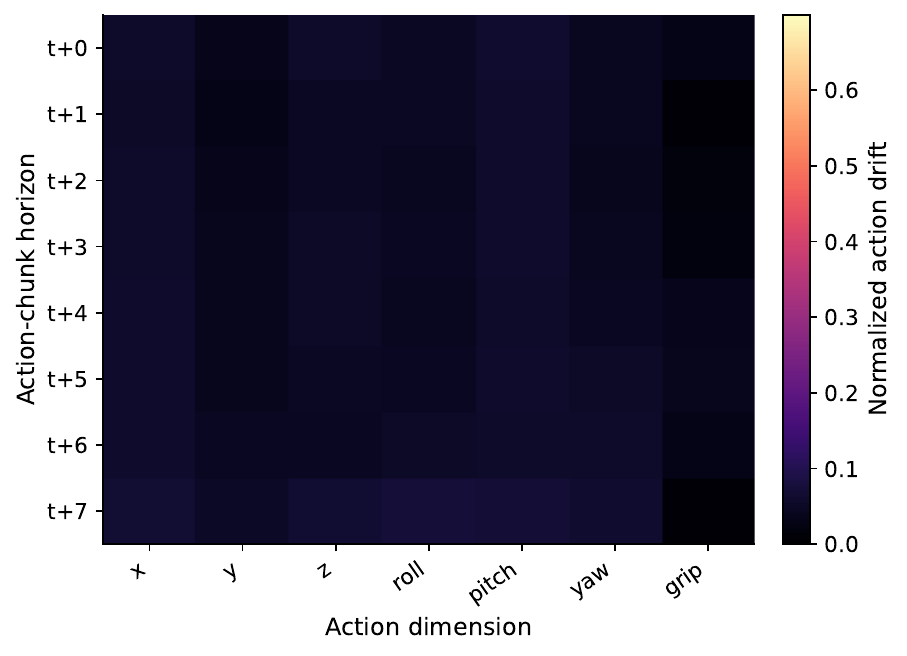}
        \captionsetup{font=footnotesize}
        \captionof{figure}{\textbf{CloudEdgeVLA action-chunk drift at $d{=}20$.}
        The shared color scale matches Figure~\ref{fig:supp_chunk_oft}.}
        \label{fig:supp_chunk_cloudedge}
    \end{minipage}
\end{figure}

\begin{figure}[H]
    \centering
    \includegraphics[width=0.78\textwidth]{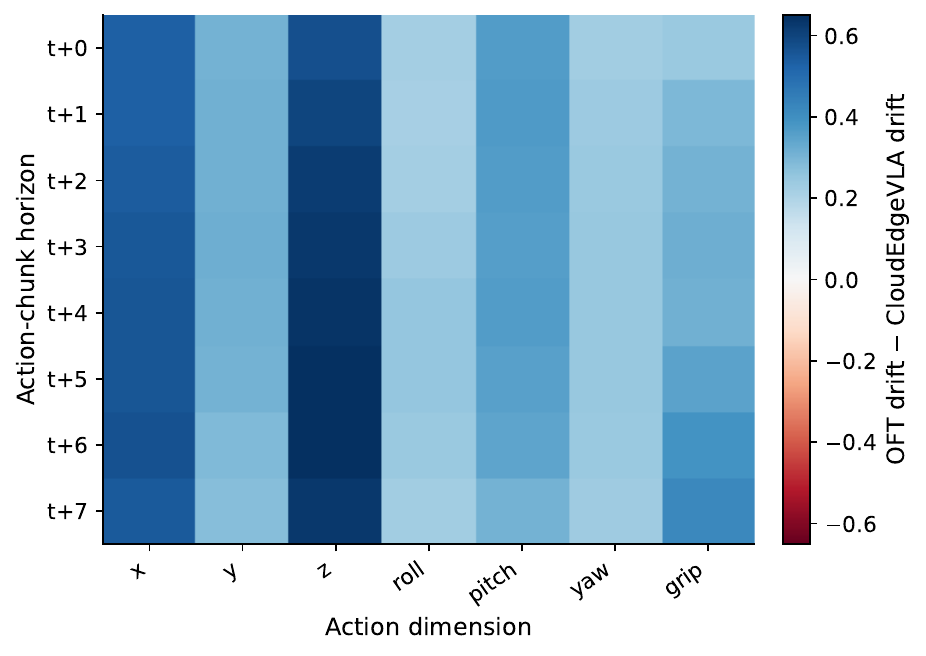}
    \captionsetup{font=small}
    \caption{\textbf{Action drift suppressed by delay training.}
    Positive values denote OpenVLA-OFT drift minus CloudEdgeVLA drift.}
    \label{fig:supp_chunk_suppression}
\end{figure}

The reduction is distributed across the complete action chunk. Comparing Figures~\ref{fig:supp_chunk_oft} and~\ref{fig:supp_chunk_cloudedge}, CloudEdgeVLA has less drift throughout the horizon. Figure~\ref{fig:supp_chunk_suppression} confirms suppression in all 56 horizon--dimension cells. The mean suppression is 0.376, with cell values ranging from 0.214 to 0.651. Translation in $x$ and $z$ shows the largest average reductions, but rotation and gripper dimensions also improve throughout the horizon.

\ifdefined\CloudEdgeAppendixBodyOnly
\let\CloudEdgeAppendixEnd\relax
\else
\def\CloudEdgeAppendixEnd{\end{document}}
\fi
\CloudEdgeAppendixEnd

\fi

\end{document}